\documentclass[letterpaper, 10 pt, conference]{ieeeconf}  

\IEEEoverridecommandlockouts  
\usepackage{cite}
\usepackage{comment}
\usepackage{amsmath, amssymb, amsfonts}
\usepackage{algorithm}
\usepackage{algpseudocode}
\algnewcommand{\To}{\textbf{To }}
\algnewcommand\Thread{\item[\textbf{Thread:}]}%
\algnewcommand\Result{\item[\textbf{Result:}]}%
\algnewcommand\Parameter{\item[\textbf{Parameters:}]}%
\algnewcommand\Data{\item[\textbf{Data:}]}%
\algnewcommand\Case{\item[\textbf{Case:}]}%
\usepackage{graphicx}
\usepackage{textcomp}
\usepackage{stfloats}
\usepackage{xcolor}
\usepackage{amssymb}
\usepackage{cite}

\def\BibTeX{{\rm B\kern-.05em{\sc i\kern-.025em b}\kern-.08em
    T\kern-.1667em\lower.7ex\hbox{E}\kern-.125emX}}
    
\begin{document}


\title{Environmental Awareness Dynamic 5G QoS for Retaining Real Time Constraints in Robotic Applications}


\author{Gerasimos Damigos$^{1*}$, Akshit Saradagi$^{2}$, Sara Sandberg$^{1}$ and \\ George Nikolakopoulos$^{2}$
\thanks{This project has received funding from the European Union’s Horizon 2020 research and innovation programme under the Marie Skłodowska-Curie grant agreement No 953454.}
\thanks{$^{1}$ The authors are with Ericsson Research, Ericsson AB, Lule\aa, Sweden.\,\,}
\thanks{$^{2}$The authors are with the Robotics and AI Group, Department of Computer, Electrical and Space Engineering, Lule\aa\,\, University of Technology, Lule\aa, Sweden.\,\,}
\thanks{Corresponding Author's Email: \texttt{gerasimos.damigos@ericsson.com}}}%
\maketitle


\begin{abstract}
The fifth generation (5G) cellular network technology is mature and increasingly utilized in many industrial and robotics applications, while an important functionality is the advanced Quality of Service (QoS) features. Despite the prevalence of 5G QoS discussions in the related literature, there is a notable absence of real-life implementations and studies concerning their application in time-critical robotics scenarios. This article considers the operation of time-critical applications for 5G-enabled unmanned aerial vehicles (UAVs) and how their operation can be improved by the possibility to dynamically switch between QoS data flows with different priorities. As such, we introduce a robotics oriented analysis on the impact of the 5G QoS functionality on the performance of 5G-enabled UAVs. Furthermore, we introduce a novel framework for the dynamic selection of distinct 5G QoS data flows that is autonomously managed by the 5G-enabled UAV. This problem is addressed in a novel feedback loop fashion utilizing a probabilistic finite state machine (PFSM). Finally, the efficacy of the proposed scheme is experimentally validated with a 5G-enabled UAV in a real-world 5G stand-alone (SA) network.

\end{abstract}


\section{Introduction}
\label{intro}

The emergence of 5G networks and the Fourth Industrial Revolution or Industry 4.0 concept is driving the transformation in manufacturing towards flexible and re-configurable systems that can adapt to the real-time demands of robotic applications and other time-critical use cases, such as the automotive industry, cellular-connected Industrial Internet of Things (IIoT) and others~\cite{5G_automotive, 5gamericas:urllc, rao2018impact}. Moreover, plenty of paradigms in the literature employ 5G and QoS features, and there have been substantial efforts in related industries and standardization bodies to provide the stringent performance that a time-critical application needs. The 3GPP New Radio (NR) specifications that describe the radio access technology of 5G have adopted ultra-reliable and low-latency communication (URLLC) features (Releases 15-17) ~\cite{3gpp:release15, 3gpp:release16, 3gpp:release17}. On this axis, many organizations started to integrate 5G networks with various robotics applications \cite{voigtlander20175g, 5G_industrial_robotics_2019} and the accompanying research followed in that end. 

In many communities, the 5G terminology is just a use of buzzwords today, but the real-life deployments are still far away. The use of URLLC is important for some use-cases, but for many other, the co-design of the application and the network can provide as good performance with the 5G features available today. For some use-cases, the priority that can be achieved through QoS features is a solution that can produce the expected outcome in Time Critical Communications (TCC) and robotics applications without requiring URLLC functionality. To this end, this study highlights the challenges of a 5G-enabled UAV (5G-UAV) and inspired by ongoing research and industry trends, highlights the main challenges that arise along with potential solutions.

The utilization of 5G networks with UAVs can be arguably divided in a few major axes. Ubiquitous communication coverage, high bandwidth, low-latency, and 5G localization are among the first headlines. Additionally, the main use-cases often refer to scenarios like enhanced teleoperation or the utilization of an edge cloud server where a robotics platform can harvest the extended capabilities of a cloud environment. There are two common denominators in all the described scenarios, that is the latency and throughput requirements. Such use-cases often require the large data transmission under low-latency conditions.

\begin{figure}[h!]
    \centering
    \includegraphics[width=1.\columnwidth]{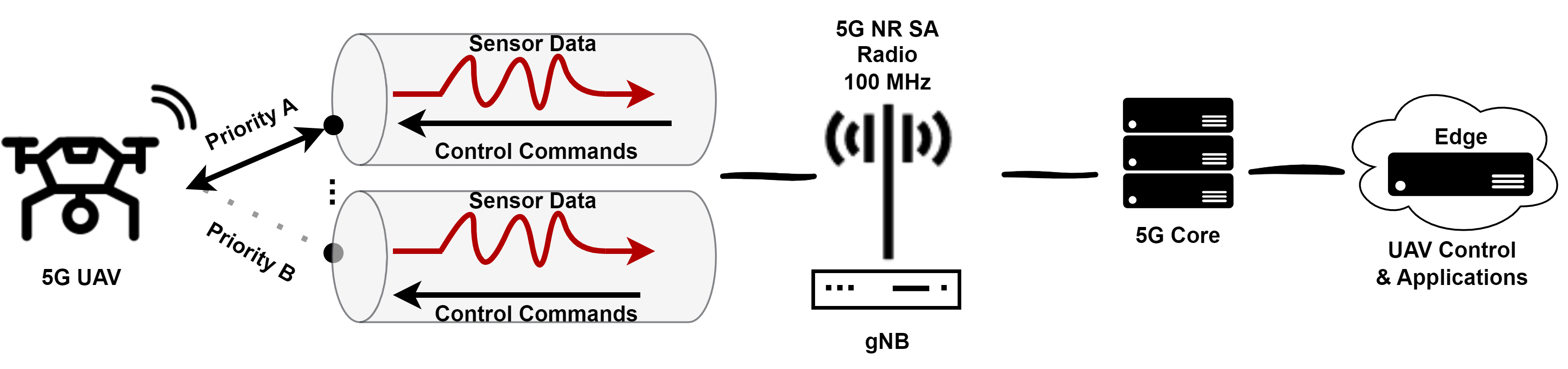}
    \setlength{\abovecaptionskip}{-10pt}
    \caption{5G-cloud-enabled UAV with distributed components in the cloud. The cloud offloading of time-critical algorithms are supported by various configurable QoS profiles.}
    \label{fig:intro-concept-fig}
\end{figure}

In this article a framework where a 5G-UAV offloads time-critical and high-throughput operations to the edge cloud is presented. As such, a model predictive control (MPC) position controller is offloaded to the edge cloud for the control of the 5G-UAV along with high bit rate camera data so that computational intensive object detection algorithms can be used in real-time. The focus of this study does not lie on the specific utilized algorithms, rather than the establishment of a 5G-UAV QoS selection framework to ensure that any similar operation can be reliably executed. More specifically, the presented framework focuses on the utilization of 5G QoS to mitigate presented latency and throughput challenges on the closed loop operation of the 5G-UAV caused by network load, while taking into account the environmental conditions that the UAV operates in, i.e., high risk in cluttered space. A high-level overview of the architecture is depicted in Fig. \ref{fig:intro-concept-fig}. 

Finally, the contributions presented in this study have been experimentally demonstrated in real-life and can be summarized as follows: 1) This research presents a novel integration of 5G networks and the dynamic use of QoS features with UAVs to maintain their real-time constraints. Note that, to the best of our knowledge, the dynamic use of QoS in 5G-UAVs has never been tested before in real-life. 2) The designed system can dynamically adjust to network load and preserve optimal performance by maintaining high throughput. 3) The proposed solution incorporates totally novel environmental awareness and fallback actions to the solution, so that the selected QoS is based on the current risk of the 5G-UAV and safety is maintained even in out of coverage scenarios. 
\subsection{Related Work}
In this Section, the state-of-the-art related work regarding the integration of time-critical robotics with 5G networks is briefly introduced. With the maturity of 5G technology, many studies started to investigate the key aspects around this cross-section, as well as potential benefits as mentioned in~\cite{zeng2019accessing, mishra2020survey, hu2021vision, voigtlander20175g}. A highlighted topic refers to the fact that robotic applications that operate over a wireless channel must include the analysis of the wireless channel~\cite{licea2023communications}, e.g., communication aware planning is essential to maintain good radio conditions~\cite{jorgensen2022towards, mardani2019communication}. Another popular topic refers to the offloading of computationally demanding applications to the edge cloud, such as optimization based controllers (e.g., MPC) or object detection algorithms. Here 5G networks are the interface that provides the high throughput data transmission and low latency communication~\cite{zhu_IROS_2023, damigos2023toward, seisa2022cnmpc, sossalla2022offloading, sankaranarayanan2023paced}. Lately~\textit{Chen et al.} \cite{Chen_IROS_teleoperation} demonstrated high dexterity teleoperation over a 5G network. However, most of these studies consider a simulated network (that usually diverges from real-life full-scale network conditions) and do not consider any of the advanced QoS features of 5G to tackle communication-related challenges. Nevertheless, recently a study in that direction emerged, where time-critical requirements for the control of a 5G-edge-robotics paradigm were fed into the 5G's scheduling algorithm to optimize network traffic and RTT latency~\cite{intel_2023_communication_control}, the validation of the solution was experimentally tested in a simulated environment. Further, another common factor for all the aforementioned solutions resides in the ability to transmit high throughput sensor data in real time and the corresponding challenges~\cite{damigos2023performance}. Finally, it's key to stress that there are a lot of current studies targeting futuristic solutions, while limited studies have actually demonstrated real-life experiments. That extends to an understudied sub-field of the described joined field, where existing technology has not yet been utilized to address real-life challenges.       

\section{System Architecture}
\label{resource_allocation}
\begin{figure*}[t!]
    \vspace{-5mm}
    \centering
    \includegraphics[width=1.\textwidth]{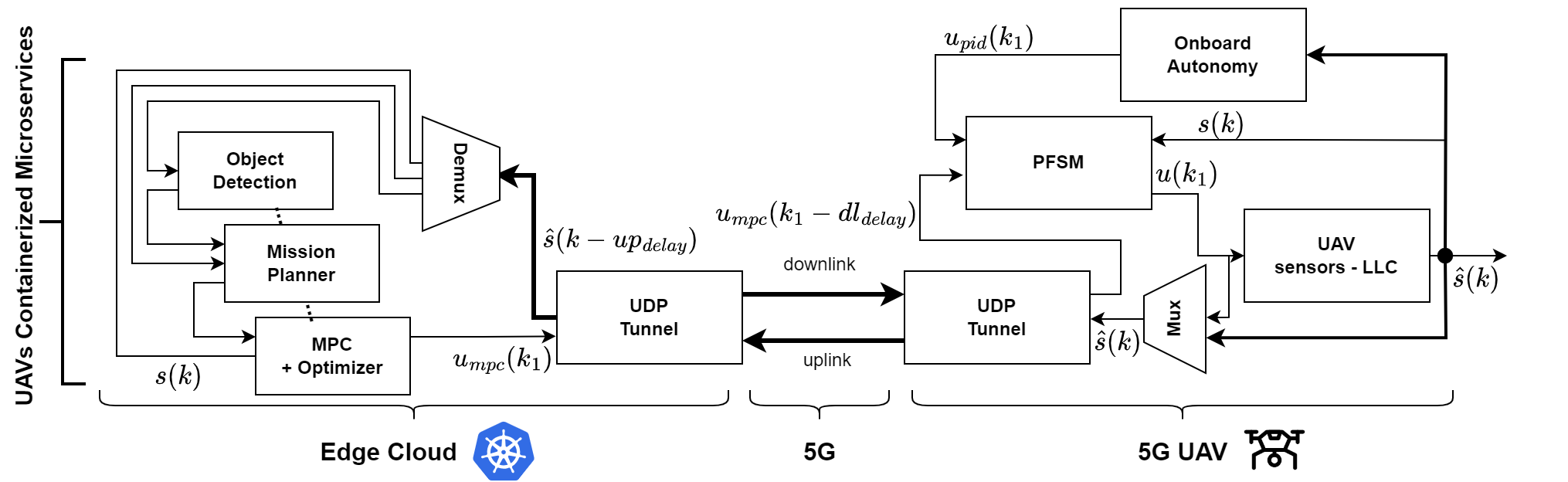}
    \setlength{\abovecaptionskip}{-10pt}
    \caption{System architecture and distributed component diagram.}
    \label{fig:methodology-architecture}
\end{figure*}

This Section presents the correlation between the 5G-UAV performance, the sensed environment and the network-related Key Performance Indicators (KPIs). An architecture similar to the one presented at~\cite{damigos2023toward} is used and visualized in Fig. \ref{fig:methodology-architecture}. The foundation of this architecture considers time-critical components, such as the position controller, object detection, and path-planning algorithms of the UAV, located offboard, i.e., in the edge cloud of the 5G network infrastructure. The most important KPI refers to the round trip time (RTT) of the control \& command data which is defined later. Further, a risk factor is calculated based on the proximity of objects to the UAV through an RGB depth sensor. 

Regarding the communication aspect, the key challenges are to maintain low latency and jitter on the closed loop architecture, all while transmitting essential high throughput sensor data in real-time. The two challenges are coupled in nature and their success relies on the radio conditions of the 5G-UAV (coverage), as well as network load conditions, e.g., significant background load. Thus, it is of utmost importance to eliminate additional substantial delays and maintain the closed-loop performance of the 5G-UAV. This study presents a novel approach where a probabilistic finite state machine (PFSM) is employed and dynamically utilizes 5G QoS data flows and potential fallback actions, while employing environmental awareness.

\subsection{Problem definition}
\label{sub:problem-definition}
The correlation between the behavior of a 5G-UAV, the expected RTT latency, and the additional buffer delay has been studied and described in the literature~\cite{damigos2023toward, wu2022modeling, sankaranarayanan2023paced}. The problem can be summarized as follows: the state of the robot $s(k)$ is captured at time $k$ and has to be transmitted to the remote controller hosted at the edge server of the network. The time that the state packets need to get into the edge server of the 5G network (for the uplink direction) is captured by the value $up_{delay}$, and the state of the robot in the edge server corresponds to the delayed state $s(k - up_{delay})$. The remote MPC controller \cite{sankaranarayanan2023paced} located in the edge server along with the mission planner (that indicates the desired path in a form of waypoints) computes an action command $u(k_1)$ at time $k_1$ and sends it to the 5G-UAV. The time that $u(k_1)$ needs to get to the UAV in the downlink direction corresponds to the value of $dl_{delay}$, and the command that reaches the 5G-UAV is $u(k_1 - dl_{delay})$. By combining the delays in the uplink and downlink direction, the RTT latency can be defined as $RTT = ul_{delay} + dl_{delay} + t_{exe} + p_{delay}$, where $t_{exe}$ refers to the execution time of the MPC controller and $p_{delay}$ refers to the processing delay in the sensor data transmission. The processing delay includes the operations of data pre-processing and post-processing, data compression, data serialization and de-serialization. The later two delay components, i.e., $t_{exe}$ and $p_{delay}$ are ignored from this analysis as they are considered independent of the network analysis. Finally, the combined state of the robot and the surrounding environment, i.e., information captured by the Intel RealSense d455 (RGB depth) sensor are denoted as the extended state and captured at the edge server as $\hat{s}(k-up_{delay} - p_{delay}/2) = \hat{s}(k-up_{delay})$.    

Multiple studies have demonstrated that an increase in RTT latency and jitter has a direct impact on the behavior of the 5G-UAV~\cite{damigos2023toward, sankaranarayanan2023paced, seisa2022edge}. As a general guideline, the RTT latency should be lower than the remote controller's operation period $T_{MPC}$, meaning that $RTT \le T_{MPC}$. If this condition is met, the operating latency of the 5G-UAV is not noticeable. To estimate the value of the RTT latency in real-world scenarios of 5G-UAVs, relying on statistics from such measurements is advisable. While some studies can provide an analytical definition of the RTT, it requires complete knowledge of the deployment, as well as real-time insights into all the user's traffic. Hence, this paper considers that the expected RTT latency that will characterize the 5G-UAV's behavior can be captured by prior network statistics~\cite{wu2022modeling, sankaranarayanan2023paced}.

The RTT statistics can vary depending on many parameters like network or cell load and various radio KPIs, like the signal to interference noise ratio (SINR) and others~\cite{damigos2023performance, sankaranarayanan2023paced}. However, one very important parameter is the amount of data being transmitted. More specifically, if the data being transmitted exceeds the uplink capacity of the 5G-UAV, then significant additional latency will be observed on-top of the expected RTT. The additional latency is an outcome of full data buffers on the 5G-UAV side and can result in instability of the closed-loop system~\cite{damigos2023toward, damigos2023performance}. For the studied scenario let the 5G-UAV have a required throughput both in the uplink and downlink directions, $th_{ul}$ and $th_{dl}$ respectively. Also, the expected sensor data bit rate produced by the 5G-UAV is denoted as $BR_i$ and considers the uplink direction. Then when $BR_i \geq th_{ul}$ the RTT will grow aggressively and the additional latency from the buffer delay can make such a system unstable \cite{sankaranarayanan2023paced, damigos2023performance}. The main motivation behind this study is to remove the additional buffer delay, thus; ensuring operation under the expected RTT value that the 5G NW can support under the current load. Here, the idea is to dynamically assign to the 5G-UAV different resources in time, so that $BR_i \leq th_{ul}$, and consequently the tracking error is not related to the additional delay of the system.

\subsection{QoS \& Dynamic QoS motivation}
In the context of 5G Quality of Service (QoS) features, priority assignment and the allocation of additional resources to specific data flows are crucial aspects. These resources play a pivotal role in meeting the real-time requirements of time-critical applications. However, it's essential to delve into the rationale behind the dynamic utilization of these resources and the reasons for not employing QoS continuously.
One might argue that time-critical applications, such as a 5G-UAV, should always be given top priority over other users in the network. The motivation behind the dynamic use of QoS stems from the possibility of other important users requiring priority as well. When multiple users already require improved QoS, their prioritization can become ineffective due to their access to the same scheduling weights. Therefore, the proposed solution adopts a need-based approach, considering both environmental awareness and an estimated latency expression for the UAV's RTT (which corresponds to tracking error in high RTT conditions). This approach aims to efficiently optimize resource allocation.    

\subsection{5G QoS \& Relative Priority}

Scheduling algorithms in 5G must allocate resources to meet specific QoS requirements, including prioritizing data flows for different applications or within the same device. For example, in scenarios involving a time-critical robot mission and a regular user streaming video, resource allocation should reflect the varying significance of these applications. Prioritization is vital to ensure that critical data flows receive the necessary resources, such as control and command data for the robot.

The prioritization between flows can be either of relative nature or absolute nature. Relative priority considers how much resources a user utilizes compared to other users, while absolute priority aims at enforcing a certain throughput for the prioritized user without taking the consequences for other users into account. This study employs a \textit{resource-fair scheduler} with relative priority QoS flows to meet the real-time requirements of a 5G-UAV. The relative priority scheme is realized in an algorithmic fashion and is established by weighting different users or data flows in time compared to others. This weight is then fed as an input parameter to the scheduler and the prioritized user is favored compared to a user with a lower relative priority \cite{scheduling-3gpp, 5G_schedulers_survey}.  

Using a weight calculation function one can calculate how the relative weight of each user increases as a function of time as long as the user is not scheduled. This description uses the term \textit{user} for simplicity, but it can equally well be data flows with different priority from the same UE. The relative weight is used by the scheduler when assigning resources and the higher the weight the more resources the user will be assigned. Once a user is assigned resources, the weight is reduced. In many cases, commonly used weighting functions employ simple line functions like $w_{i}[k] = R_i \cdot k$, where $R_i$ depicts the relative priority parameter, in this case the slope of the function, and $k$ the discrete time. The slope ratio for two different users establishes the relative priority ratio between them. For example, $w_{user_1}[T] = 2 \cdot w_{user_2}[T]$, or $w_{user_1}[T/2] = w_{user_2}[T]$. However this ratio does not directly correspond to a \textit{2:1} resource allocation to the users. The final resource allocation that the scheduling algorithm assigns is subject to other essential parameters such as the amount of data in each user's buffer, i.e., how much data the user wishes to transmit, and the traffic pattern. Additionally, note that the resource assignment with a resource-fair scheduler is independent of the channel conditions and potential interference. Nevertheless, channel conditions and interference define the user's radio conditions and thus communication capabilities (e.g. throughput and latency). Thus, in order for one to observe the expected throughput behavior, e.g., a \textit{2:1} throughput ratio, the two users must exist under the same radio conditions. 

This study presents results and experimental evaluations that consider users of equivalent radio conditions, while additional studies are planned to examine varying radio conditions. Such a separation in scenarios makes the study robust.

\subsection{Dynamic Resource Selection}
\begin{figure}[h!]
    \centering
    \includegraphics[width=.9\columnwidth]{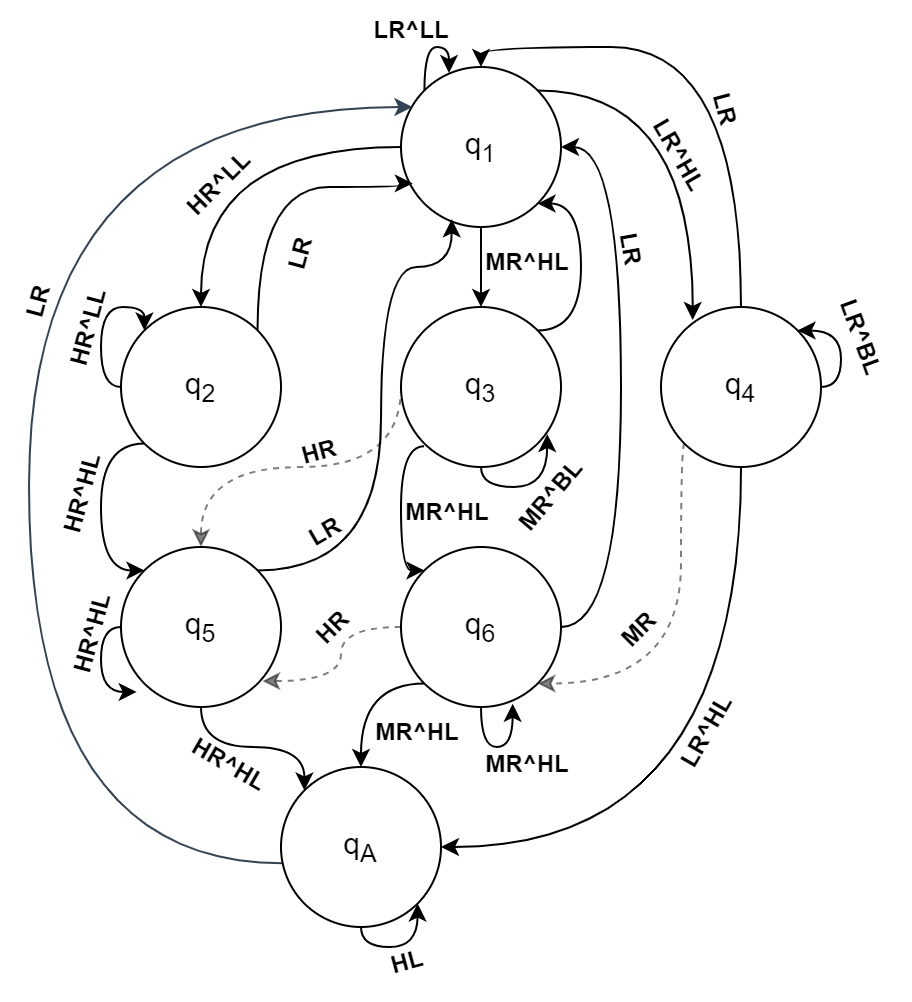}
    \caption{PFSM architecture.}
    \label{fig:state-machine}
\end{figure}

A Probabilistic Finite State Machine (PFSM) solution is utilized for the demonstration of the proof of concept for dynamic QoS selection paradigm. The primary objective of the PFSM is to facilitate seamless transitions between default data flows, prioritized data flows, and full onboard autonomy fallback mode, all while implementing data rate adaptation mechanisms (on the application side) to ensure the desired behavior of the 5G-UAV. The most significant challenge addressed in this study pertains to the mitigation of substantial buffer delays, which can render the system unstable. Additionally, environmental awareness is incorporated into the system. The complete diagram of the utilized state machine is presented in Fig. \ref{fig:state-machine} for reference. \\ 

\textbf{Definition 1:} The presented is variation of the utilized PFSM is inspired by other studies \cite{allison2017uav} and is a tuple 
$M = <\Sigma, Q, P, \delta, \theta>$, where
\begin{itemize}
    \item[-] $\Sigma$ is the input alphabet based on environmental stimulus.
    \item[-] $Q={q_1, q_2, ..., q_6}$ is finite set of states  
    \item[-] $\delta: Q \times \Sigma \times Q \rightarrow \mathbb{R} \cap [0,1]$ the set of transitions
    \item[-] $P$ is the set of transition probability functions that map to $\delta$
    \item[-] $\theta$ is a set of response thresholds
\end{itemize}

The PFSM design is based on the combination of 6 signals and 7 states, the definition of each state is found in table \ref{tab:state-table}, and its operation is executed in a feedback fashion. A synopsis behind the intuition of the signals is that they are a logical combination of two things, a \textit{risk factor}, i.e., low, medium, and high risk, and similarly, the communication RTT latency, more specifically, low and high RTT latency. The risk factor is based on a quantitative estimation of cluttered space at a proximity threshold near the 5G-UAV and is inspired by \cite{chen2022direct}. The states include a combination of QoS-enabled prioritized data flows, default data flows, congestion mechanisms for large data transmission, and full onboard autonomy. The utilized threshold functions are based on a logistic sigmoid form and the response thresholds are defined by the transition probability
\begin{equation} \label{eq:sigmoid}
    P(t_i) = \frac{1}{1 + e^{\mu_i (\rho_i - t_i)}}
\end{equation}
where $t_i$ represents the magnitude of a stimulus $t$ for a specific parameter $i$, i.e., latency or risk. $\rho$ represents the offset along the x-axis (essentially the threshold) and $\mu$ is the slope of the curve. Equation \ref{eq:sigmoid} is then utilized along with a corresponding threshold $th_j$ to produce the actual conditions that generate the transition signals, i.e., $HR$, $LL$, and so on. The sigmoid functions are used in 3 ways, for the camera data latency, for the control data latency, and finally the risk factor. More specifically, the latency condition is a weighted sum of the two mentioned latency sigmoids and is defined as follows
\begin{equation} \label{eq:latency-cond}
        P_{lat}(t_1, t_2) = w_1 \cdot P_{cam}(t_1) + w_2 \cdot P_{cc}(t_2), \\
\end{equation}
where $w_1 + w_2 = 1$, while $P_{lat}$, $P_{cam}$, and $P_{cc}$ are the transition probabilities for the joined latency condition, the camera data latency and the control $\&$ command RTT latency, respectively. Then, $t_1$ and $t_2$ are derived by a sliding window average of the $N$ last latency samples as
\begin{align} \label{eq:sliding-window}
    &\ t_i = \frac{1}{N} \sum_{j = k - N}^{k} delay_i(j), \nonumber \\ 
    &\ where \; i \in [1,2] \in \mathbb{Z} \, and \, j \in [1, +\infty] \in \mathbb{Z}  
\end{align}
where, $i = 1$ represents the camera data, $i = 2$, the control data of the 5G-UAV, and $k$ is the discrete time. In a similar fashion, the risk condition is directly derived by Eq. \eqref{eq:sigmoid} and defined by the transition probability for identifying cluttered space in proximity, $P_{cs}(t3) = P(t_3)$. In that case $t_3$ expresses the spaciousness of the sensed space in proximity and is derived on the point cloud of the RGBd sensor as follows: $s_k = \alpha \cdot s_{k-1} + \beta \cdot S_{k}$, where $s_k$ stands for the current sensed spaciousness at time $k$ and $S_k$ is determined by calculating the mean Euclidean distance from the center mass of the drone to each point of the sensed point cloud. For example, the considered value of $s_k \leq 3\;m$ is defined and yields a high risk condition for the 5G-UAV (HR signal in the state machine), and thus would require the maximum amount of allocated resources to be assigned proactively even without increased latency. In addition, the middle risk environmental condition is defined in the case that $3 < s_k \leq 5$ and activates the middle risk signal. 

Finally the states can be divided in three categories. The default QoS category $q_1$ and $q_4$, where no QoS priority is applied and a low risk is detected, the QoS enabled category, $q_2$, $q_3$, $q_5$ and $q_6$, where $HR$ or $MR$ is detected, and the full onboard autonomy category, $q_A$ where the UAV does not require any external communication and operates with the onboard autonomy stack. The actions of the said states are explained in table \ref{tab:state-table}. Note that the full autonomy solution is inspired by \cite{damigos2023resilient} and includes a fallback onboard PID position controller and an onboard planner, whereas the rate adaptation action periodically reduces the camera data rate and seeks to avoid the rising buffer delay by sacrificing throughput. This action occurs when the QoS priority has already been applied but the buffer delay still rises (e.g., multiple prioritized users fight for resources). Note that there are state-of-art rate adaptation mechanisms \cite{l4s_ieee, ericsson-whitepaper}, but there are not the focus of this study. The proposed solution was selected to mitigate integration challenges. 

\begin{table}[h]
\centering
\begin{tabular}{|c|c|c|}
\hline
Current State & Action & Next States \\
\hline
$q_1$ & Offload - no QoS & $q_1, q_2, q_3, q_4$ \\
\hline
$q_2$ & Offload - QoS & $q_1, q_2, q_5$ \\
\hline
$q_3$ & Offload - QoS & $q_1, q_3, q_5, q_6$ \\
\hline
$q_4$ & Offload - no QoS - rate adaptation & $q_1, q_4, q_6, q_A$ \\
\hline
$q_5$ & Offload - QoS - rate adaptation & $q_5, q_1, q_A$ \\
\hline
$q_6$ & Offload - QoS - rate adaptation & $q_6, q_1, q_A$ \\
\hline
$q_7$ & Full onboard autonomy & $q_A, q_1$ \\
\hline
\end{tabular}
\caption{The PFSM state table describes the executed actions of each state.}
\label{tab:state-table}
\setlength{\abovecaptionskip}{-10pt}
\vspace{-8mm}
\end{table}


\section{Experimental Evaluation}
\label{experiments}
For the evaluation of this study a real 5G network with the described QoS features was utilized along with an in-house built quad-rotor equipped with a 5G modem (Sierra Wireless EM9191). A vicon motion tracking system was used to provide ground truth odometry and capture the UAV's state. The 5G radio operates in mid-band frequency (3.7 $GHz$) and provides an indoor 5G Ericsson DOT base station. The edge server is located in the local breakout of the 5G system in a distance of approximately $4 \; km$ round trip. Further, the edge server hosts a Kubernetes (K8s) system that is utilized to execute all the edge-hosted components of the 5G-UAV. The maximum theoretical cell throughput of the system is approximately $94 \, Mbps$ in uplink and $1400 \, Mbps$ in downlink under good radio conditions \cite{damigos2023performance}. 

\begin{table}[h]
\vspace{-2mm}
\centering
\begin{tabular}{|c|c|c|c|}
\hline
Scenario & UAV Thr (Mbps). & RTT (msec) & BG Thr. \\
\hline
 & $Tx: \sim 47 $ &  & \\
No Qos - No BG & $Rx: \sim 47 $ & $\sim 27.3 $ & No BG Load \\
\hline
 & $Tx: \sim 47 $ &  & $Tx: \sim 80 $\\
 No Qos - BG & $Rx: \sim 40 $ & (rises) & $Rx: \sim 40 $\\
 \hline
 & $Tx: \sim 47 $ &  & $Tx: \sim 80 $\\
Priority Qos - BG & $Rx: \sim 47 $ & $\sim 28.1 $ & $Rx: \sim 32.5 $\\
 \hline
\end{tabular}
\caption{Quantitative averages of control \& command RTT latency, transmitted (Tx) and received (Rx) sensor data throughput (5G-UAV Thr.) and transmitted and received background traffic throughput (BG Thr.) for the three main scenarios. The averages are taken across 14 conducted experiments.}
\label{tab:qos-results}
\setlength{\abovecaptionskip}{-10pt}
\vspace{-5mm}
\end{table}

The proposed solution of dynamically selecting QoS profiles to ensure the safe operation of a 5G-UAV is tested as follows. The position control of the 5G-UAV is offloaded to the edge cloud along with an object detection algorithm that detects circular objects. The two applications are hosted in different K8s pods. The data needed to execute the said components is transmitted in the uplink direction from the 5G-UAV at an average throughput of $~47 Mbps$. This includes the captured state of the UAV and various camera data topics (high resolution RGB image and depth information). Note that this total occupies about $50 \, \%$ of the practical maximum uplink cell throughput. Then, at one point, a significant background traffic of $80 \, Mbps$ is initiated by another user in the same cell and the system is tested, with and without the proposed solution. Note that the total load exceeds the cell throughput with $~35 \, \%$. The parameters used for the manifestation of the experiments are the following: $[\mu_1, \mu_2, \mu_3]=[3, 5, 5]$, $[\rho_1, \rho_2, \rho_3]=[61, 27, 3]$, $[w_1, w_2]=[0.35, 0.65]$ and $[N_1, N_2]=[10, 50]$.       

\begin{figure}[h!]
    \vspace{-3mm}
    \centering
    \includegraphics[width=1.11\linewidth]{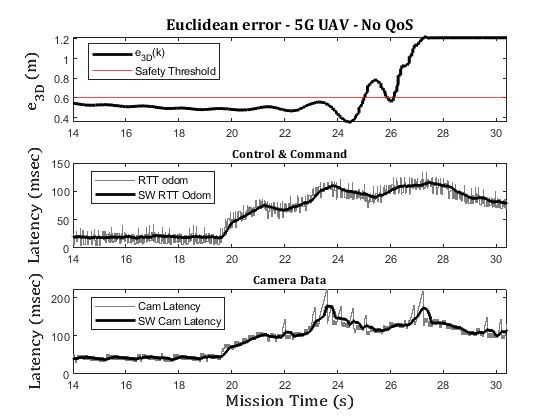}
    \setlength{\abovecaptionskip}{-10pt}
    \caption{No QoS scenario and PFSM. The 5G-UAV performs a stationary hover where the offloaded MPC and an object detection algorithm are hosted in the edge cloud. Significant background load is introduced in the network cell and the system becomes unstable.}
    \label{fig:res-no_QoS_error}
    \vspace{-5mm}
\end{figure}

Figure \ref{fig:res-no_QoS_error} illustrates a scenario in which Quality of Service (QoS) was not employed, neither in a static nor dynamic manner. In this particular scenario, the objective of the UAV mission was to maintain a stationary hover. When competing traffic is initiated, the anticipated behavior is observed: the resources are equally split resulting in a significant queue build-up in the 5G-UAV buffer (as mentioned in \ref{sub:problem-definition}); uncontrolled latency in both camera data and control \& command data rises aggressively and finally instability occurs. Essentially, the system transforms into an open-loop configuration. In this scenario, a stationary hover was chosen to avoid the damage of the robotic platform. Note that even though the theoretical maximum uplink throughput is $94 \, Mbps$, the one tested before the manifestation of the experiments yields an average value of $81.3 \, Mbps$; making practically impossible to handle that many data without filling up the buffers in both users. Finally, visual effects like video frame loss effects (i.e., lag and freeze) are also visible from the UAV's point of view and depicted in the accompanying video.  

\begin{figure}[h!]
    \vspace{-5mm}
    \centering
    \includegraphics[width=1.10\linewidth]{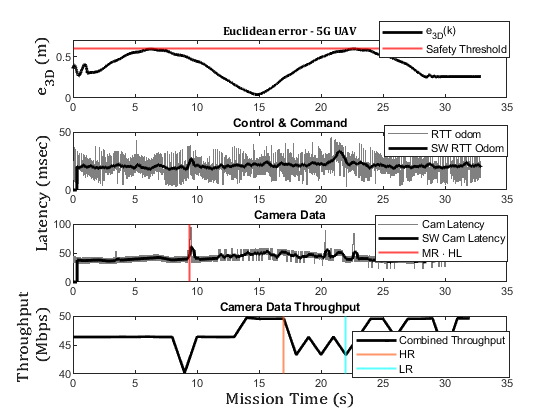}
    \caption{PFSM and dynamic QoS utilization enabled. The 5G-UAV performs a circular trajectory where the offloaded MPC and an object detection algorithm are hosted in the edge cloud. Significant background load is introduced in the network cell. The PFSM alternates between the designed states so that the 5G-UAV maintains optimal behavior and stability.}
    \label{fig:res-QoS_error}
\end{figure}

Figure \ref{fig:res-QoS_error} depicts the complete solution with the exact same load applied in the network cell as an external "disturbance". This time, the examined scenario involves a mission where the 5G-UAV takes off and has to track a circular trajectory while capturing sensor data from the RGBd camera in real-time. The circle imitates a scenario where the 5G-UAV has to execute maneuvers that are part of a full mission and the real-time camera data transmission imitates a scenario where sensor data are meant to be used along with a control, planning or a perception algorithm; thus the camera data transmission latency has to remain unaffected by the network load. On this occasion, the 5G-UAV has the ability to dynamically access the 5G QoS prioritization. When the background load is applied the transition signal of $HL$ is triggered and the risk signal is already in $MR$. Hence a transition from state $q1$ to state $q3$ occurs and the QoS prioritization is applied on the 5G-UAV's complete traffic profile. The improvement on the latency and throughput is almost immediate and the UAV's tracking error becomes in practice unnoticeable as the time resolution that the scheduler operates with is significantly finer than the one of the data transmission and the MPC's control loop. In addition, table \ref{tab:qos-results} depicts average values of the discussed KPIs among all the performed experiments.  

Ultimately its important to highlight a few takeaways. This proposed solution seeks to demonstrate the potential of the 5G QoS features and how a 5G-UAV can dynamically utilize the network resources to sustain a stable operation and enjoy the computational capabilities of an edge cloud server. This solution should be coupled with environmental awareness so any UAV in a low risk situation switch back to a data flow with default priority and increase the availability of resources for users in high risk situations. Finally, as already demonstrated, fallback solutions must be implemented. As mentioned before, resources are shared among users, thus if all users were to utilize equal elevated priority, then the concept of priority is potentially non-applicable. The later indicates potential directions in centralized solutions.


\section{Conclusions and Future Developments}
\label{conclusions}
In this article, the novel utilization of 5G QoS for retaining the system's real-time constraints was proposed and experimentally validated in real-life conditions for the first time. The outcome is clear and it showcases that the intersection of 5G-enabled cloud robotics is maturing. Nevertheless, many challenges regarding the time-critical requirements of such architectures are yet to be solved and many times the problem is assigned to the other field. Integration of such distributed systems is challenging and requires rigorous analysis and design. Moreover, the developing process is not a trivial task. However, there is plenty of available technology to be utilized and numerous extensions that the robotics and communication co-design can offer to this field. For example, the current framework is planned to be expanded in many different directions such as the introduction of a centralized multi-agent solution, where centralized awareness will make it possible to plan the tasks of the different agents in a mission according to the availability of network resources; all while including the thorough analysis of radio conditions.      

\bibliographystyle{./IEEEtranBST/IEEEtran}
\bibliography{./IEEEtranBST/IEEEabrv,references}

\end{document}